\documentclass[letterpaper, 10 pt, conference]{ieeeconf}  

\IEEEoverridecommandlockouts                              

\usepackage{amsmath} 
\usepackage{amssymb}  
\usepackage{graphicx} 
\usepackage{tabularray}
\usepackage[hidelinks]{hyperref} 
\usepackage{booktabs}
\usepackage{cite}

\usepackage[ruled,vlined]{algorithm2e}

\title{\LARGE \bf
LAGO Policy: Latency-Aware Asynchronous Diffusion Policies with Goal-Directed Collision-Free Planning for Smooth Manipulation
}

\author{Guowei Shi$^{1,2,4,*}$, Xupeng Xie$^{1,2,4,*}$, Yiming Luo$^{3}$, Jian Guo$^{2}$, Jun Ma$^{1}$, Boyu Zhou$^{4,\dag}$
\thanks{$^{*}$~These authors contributed equally to this work.}
\thanks{$^{\dag}$~Corresponding Author}
\thanks{$^{1}$~The Hong Kong University of Science and Technology (Guangzhou)}
\thanks{$^{2}$~International Digital Economy Academy}
\thanks{$^{3}$~The University of Hong Kong}
\thanks{$^{4}$~Southern University of Science and Technology}
}

\begin{document}

\setlength{\textfloatsep}{8pt plus 2pt minus 2pt} 
\setlength{\intextsep}{8pt plus 2pt minus 2pt}    
\setlength{\abovecaptionskip}{0pt}               
\setlength{\belowcaptionskip}{0pt}               

\addtolength{\jot}{-0.0pt} 
\setlength{\abovedisplayskip}{6pt} 
\setlength{\belowdisplayskip}{6pt} 
\setlength{\abovedisplayshortskip}{6pt} 
\setlength{\belowdisplayshortskip}{6pt} 

\bstctlcite{BSTcontrol}

\maketitle
\thispagestyle{empty}
\pagestyle{empty}

\begin{abstract}
Diffusion-based visuomotor policies deployed with asynchronous inference often exhibit inter-chunk discontinuities and lack explicit mechanisms for obstacle-aware execution, leading to jerky motions and collisions that hinder reliable manipulation in real-world scenes. To address these issues, we propose LAGO Policy, a unified asynchronous action-generation framework that integrates trajectory optimization with diffusion policy for smooth and safe execution. LAGO Policy improves inter-chunk consistency via latency-aware classifier-free guidance conditioning on future actions. It further enables goal-directed collision-free trajectory planning by predicting a task-relevant interaction goal from demonstrations. Finally, spatial-temporal trajectory optimization refines the actions to be executed for low-jerk and feasible motion. Extensive real-world experiments demonstrate that LAGO Policy achieves smooth collision-free execution with high task success across challenging manipulation tasks. 
\texttt{https://lago-policy.github.io/}

\end{abstract}


\section{Introduction}

Recently, visuomotor imitation learning has emerged as a scalable paradigm for acquiring robotic manipulation skills directly from expert demonstrations~\cite{zhao2023learning, pan2026adonoisingdispellingmyths, ze20243ddiffusionpolicygeneralizable,wang2025HDP}. 
In this context, generative control policies (GCPs) instantiate closed-loop visuomotor controllers via generative models, mapping high-dimensional observations to multi-step action sequences. 
In particular, diffusion-based policies generate actions through iterative denoising and have demonstrated strong performance across a broad range of manipulation tasks.
Despite these advances, deploying diffusion-based GCPs on physical manipulators still faces fundamental challenges.
 
Firstly, ensuring continuous and smooth motion under non-negligible inference latency remains unresolved~\cite{Sigmund2025streaming}.  
To avoid stop-and-infer execution, policies are typically deployed with asynchronous inference, where the next action chunk is generated in parallel while the robot executes the current one.
This creates a perception-execution misalignment because each chunk is conditioned on observations acquired before inference, but is executed after inference finishes, when the robot and scene states may have changed. 
Consequently, consecutive chunks can disagree over their overlap, producing jerky motions that degrade fine manipulation.
Existing action-conditioning methods use predicted future actions to improve inter-chunk consistency, but tightly coupling them with observations makes the policy brittle to temporally shifted future-action conditions and prone to discontinuities at chunk boundaries~\cite{Arachchige2025SAIL}.

Secondly, GCPs lack an efficient mechanism for safe action generation in the presence of unseen obstacles. 
Since most GCPs are trained to imitate demonstrations without explicitly modeling the collision-free feasible set, they can generate geometrically infeasible trajectories once out-of-distribution obstacles obstruct the motion toward the task goal, especially during long-range goal-reaching motions.
Existing inference-time safeguards, such as collision-cost guidance or safety filters, are inherently local and short-horizon, often yielding suboptimal avoidance behaviors.
Moreover, they can push execution away from the demonstration manifold, causing safety-induced distribution shift and compounding errors under closed-loop execution.

\begin{figure}
    \vspace{3mm}
	\centering
        \includegraphics[width=0.97\linewidth]{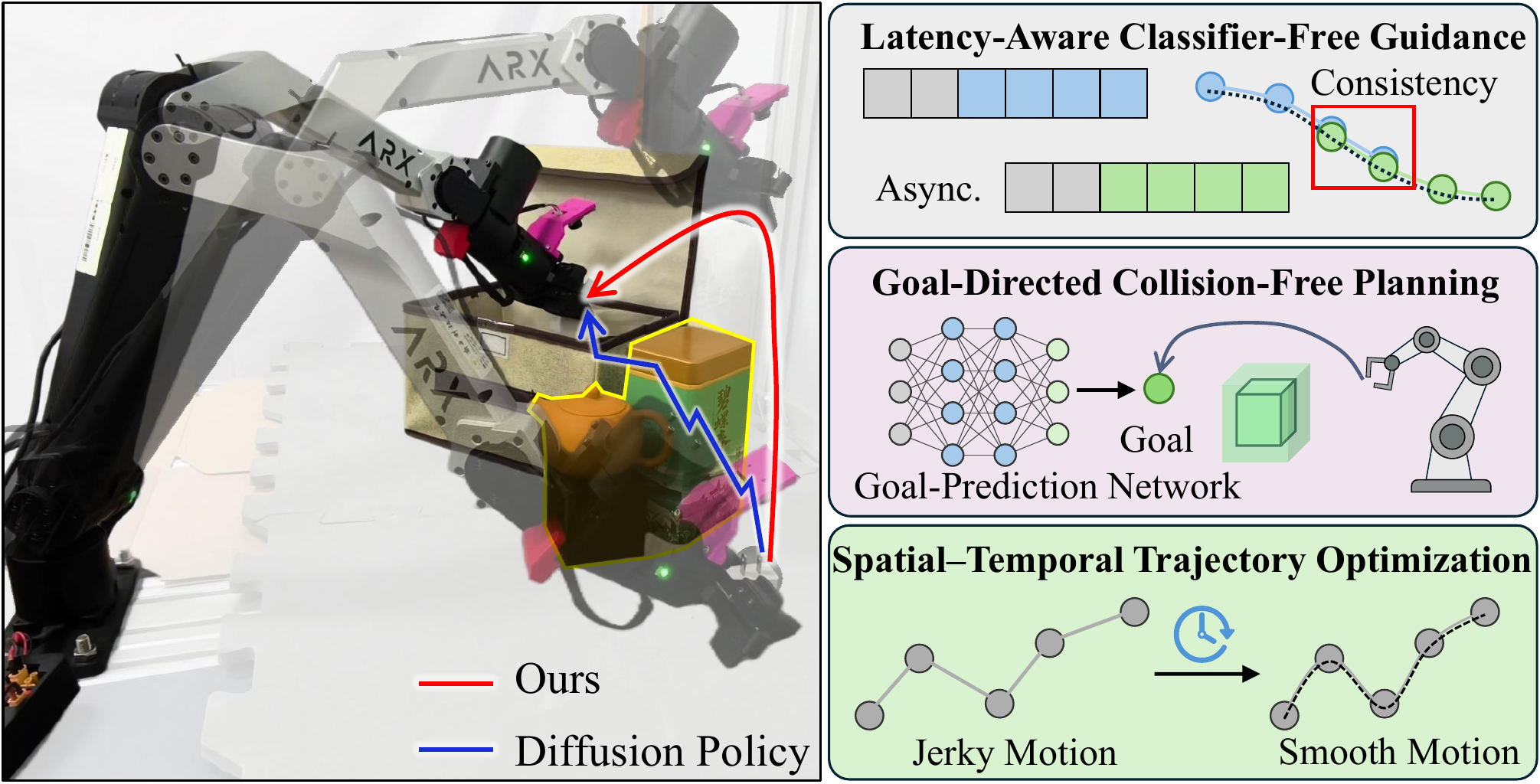}	
        \caption{LAGO Policy. Latency-aware classifier-free guidance improves inter-chunk consistency under asynchronous inference, goal prediction enables goal-directed collision-free planning, and spatial-temporal trajectory optimization refines the executed motion for smooth execution.}
	    \label{fig:intro}
\end{figure}

To tackle these challenges, we propose LAGO Policy, a unified asynchronous action-generation framework that integrates model-based trajectory optimization with diffusion-based GCPs to achieve smooth, temporally consistent, and safe execution in real-world manipulation.
For smooth and temporally consistent execution, we introduce a latency-aware training scheme with classifier-free guidance (CFG)~\cite{ho2022classifier}. 
Specifically, we decouple the future-action condition from observation features and randomize the temporal delay of the future-action condition during training, which makes the policy robust to temporal shifts in conditioning signals. 
For safe action generation, we propose a goal-directed paradigm that jointly predicts an action chunk and a task-relevant goal learned from expert demonstrations. 
When the end-effector is far from the predicted goal or the direct motion toward it is obstructed by obstacles, we activate goal-conditioned trajectory optimization to generate a smooth collision-free motion toward the predicted goal.
This planned motion steers execution back toward in-distribution states, thereby enabling task completion.
We further apply spatial-temporal trajectory optimization to refine the executed motion, including both policy-generated action chunks and planner-generated trajectories, for smooth and time-efficient execution.
We validate our method through real-world manipulation tasks, demonstrating smooth and safe action generation while improving task success rates.
Our contributions are as follows:
\begin{enumerate}
    \item A latency-aware training scheme leveraging CFG that decouples observation and future-action conditioning with randomized delay, enabling smooth asynchronous chunked execution.
    \item A goal-directed safe generation paradigm that couples demonstration-derived goal prediction with trajectory optimization to produce safe and smooth motions.
    \item Comprehensive real-world validation demonstrating improved smoothness, safety, and task success. The code will be made public.
\end{enumerate}
\section{Related Works}
\subsection{Diffusion Models for Robotics}
Diffusion models were originally developed for high-fidelity image generation~\cite{ho2020denoising}.
More recently, diffusion models have been adopted for planning and policy learning, where their ability to represent multimodal action distributions, scale to high-dimensional action spaces, and train stably is particularly attractive.
In motion planning, Diffuser formulates trajectory synthesis as an iterative denoising process that generates complete trajectories~\cite{janner2022Diffuser}.
In imitation learning, Diffusion Policy models a closed-loop visuomotor controller via conditional action diffusion and executes in a receding-horizon manner to preserve reactivity~\cite{chi2025diffusion}.
However, this closed-loop, chunked execution also exposes a practical deployment bottleneck.
Multi-step denoising incurs non-negligible inference latency, and synchronous execution often induces stop-and-infer behavior with pauses between consecutive action chunks.
Building on Diffusion Policy, our work targets continuous real-robot execution under such latency via asynchronous inference.

\subsection{Asynchronous Inference}
Asynchronous inference enables continuous execution for diffusion-based policies by generating the next action chunk while executing the current one. 
However, the resulting prediction-execution misalignment often breaks inter-chunk consistency and induces jerky motions.
Several recent works mitigate this issue via additional conditioning or correction mechanisms.
SAIL improves continuity with CFG-style future-action conditioning~\cite{Arachchige2025SAIL}, but its observation-concatenated implementation is brittle to future-action condition shifts.
VLASH makes the policy future-state-aware by rolling the robot state forward with committed actions and conditioning on the estimated execution-time state, thereby reducing state staleness under asynchronous inference~\cite{tang2025vlash}.
However, this mechanism does not explicitly promote consistency in the overlapping segment between consecutive action chunks.
Training-time RTC improves continuity via inpainting-style action conditioning, but it is tailored to diffusion-transformer-like architectures~\cite{Peebles2023SDM} and less directly compatible with CNN-based denoisers that use a globally injected diffusion timestep~\cite{black2025trainingtimeactionconditioningefficient}.
Our work follows CFG-style conditioning and explicitly addresses future-action condition shifts under asynchronous inference.

\subsection{Collision-Free Execution for Learned Policies}
Collision-free execution of learned manipulation policies is often achieved via inference-time safeguards that modify nominal policy outputs to avoid collisions.
Cost-guided diffusion methods inject collision-related costs into the denoising process to bias sampling toward collision-free trajectories~\cite{li2025language_guided_collision}.
VLSA integrates control barrier functions as a safety layer to enforce explicit state constraints~\cite{hu2025vlsa}.
RAIL applies a reachability-based safety filter that validates candidate motions and switches to a safe fallback when violations are detected~\cite{jung2025rail}.
A common thread across these approaches is that collision avoidance is handled via local, short-horizon corrections at test time, which can yield suboptimal motions, introduce frequent interventions, and push execution toward states that are rare in the demonstrations, degrading closed-loop task performance.
In contrast, our method performs goal-directed collision-free planning toward a demonstration-derived interaction goal, producing globally consistent avoidance motion and steering execution back toward in-distribution states to support task completion.
\section{Method}
\label{sec:method}
\begin{figure*}[t]
  \vspace{3mm}
  \centering
  \includegraphics[width=0.97\textwidth]{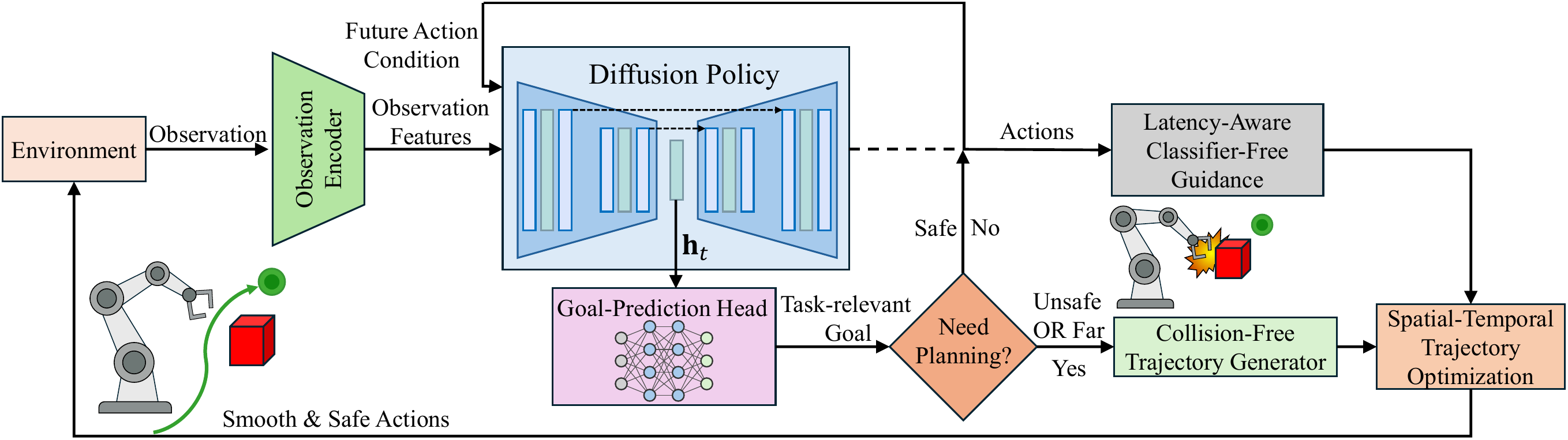}
  \vspace{-1mm}
  \caption{Overview of LAGO Policy. LAGO Policy unifies temporally consistent action generation with goal-directed collision-free motion generation for smooth robotic manipulation. Latency-aware classifier-free guidance improves inter-chunk consistency, goal prediction enables task-directed planning, and spatial-temporal optimization refines the resulting motion for low-jerk, feasible execution.}
  \label{fig:overview}
\end{figure*}

\begin{figure*}[t]
  \centering
  \includegraphics[width=0.97\textwidth]{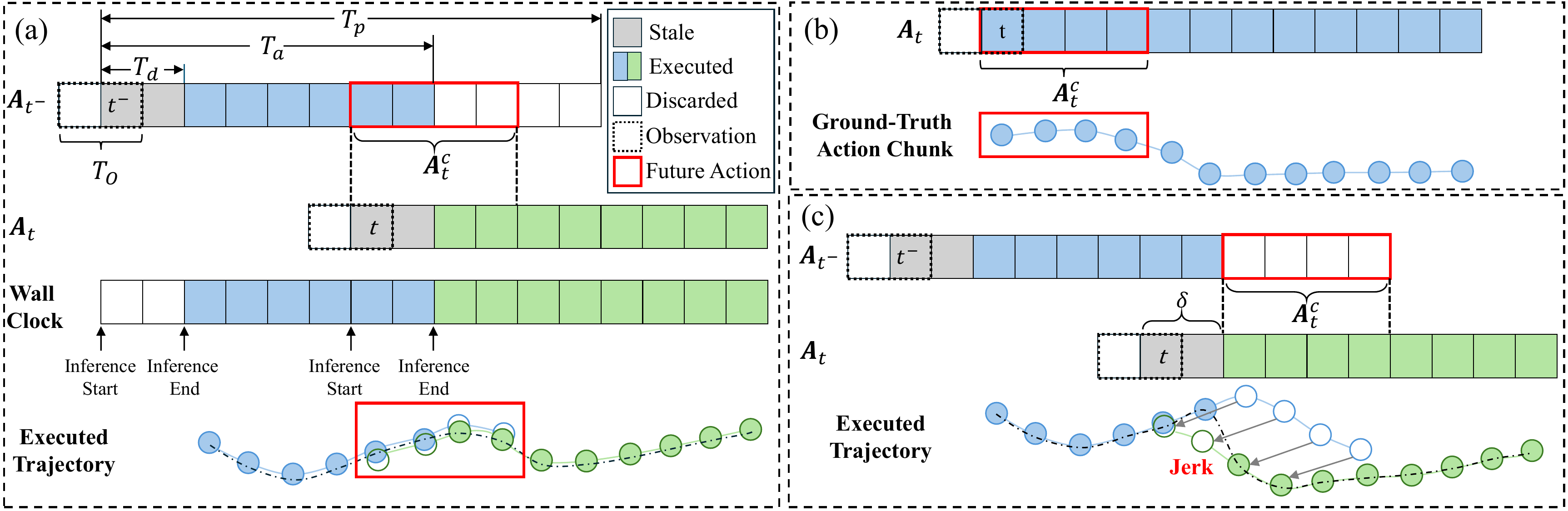}
  \caption{Motivation for latency-aware CFG under asynchronous inference.
  This figure highlights a key issue of future-action-conditioned CFG: while it improves cross-chunk continuity under temporally aligned future-action conditions, inference latency can shift the effective condition at deployment and break chunk-to-chunk smoothness.
  (a) Aligned case. Under asynchronous prediction and execution, CFG conditions denoising on the future-action condition $\mathbf{A}^{c}_t$ extracted from the unexecuted part of the previous chunk (red box), improving consistency in the overlap and smoothing the executed trajectory (black dashed line). 
  (b) Training with aligned future-action conditions. $\mathbf{A}_t$ and $\mathbf{A}^{c}_t$ are sampled from the same demonstration chunk and aligned to start at $t$ (i.e., $\delta=0$). 
  (c) Shifted future-action conditions at deployment. Inference latency shifts the effective future-action condition by $\delta$ steps, misaligning the intended overlap and inducing discontinuities and jerk motions at chunk boundaries.}
  \label{fig:train}
\end{figure*}
Fig.~\ref{fig:overview} overviews LAGO Policy.
At each asynchronous cycle, a diffusion policy with latency-aware CFG generates an action chunk, while a goal-prediction head estimates a task-relevant goal.
We trigger goal-conditioned trajectory generation when the direct motion to the predicted goal is obstructed by obstacles or the end-effector is far from the goal; otherwise we execute the policy output.
All executed motions are further refined by spatial-temporal trajectory optimization for smooth and feasible execution.

\subsection{Preliminaries}
\label{sec:Preliminaries}
Our method is built on a CNN-based Diffusion Policy backbone~\cite{chi2025diffusion}.
It models the conditional distribution
$ p_\theta(\mathbf{A}_t \mid \mathbf{O}_t)$
over a $T_p$-step future action sequence
$\mathbf{A}_t=\{\mathbf{a}_{t+k}\}_{k=0}^{T_p-1}$,
conditioned on the most recent $T_o$ observations
$ \mathbf{O}_t = [\mathbf{o}_{t-T_o+1},\ldots,\mathbf{o}_t] $,
where each observation decomposes as
$\mathbf{o}_t=[\mathbf{v}_t,\mathbf{x}_t]$ with visual input $\mathbf{v}_t$ and robot state $\mathbf{x}_t$.
The visual input is encoded as $\mathbf{z}_t=f_{\mathrm{enc}}(\mathbf{v}_t)$, and we form the per-timestep conditioning vector
$\mathbf{c}_t = [\mathbf{z}_t, \mathbf{x}_t]$. The denoising network is conditioned on the history conditioning vector
$\mathbf{C}_t=[\mathbf{c}_{t-T_o+1},\ldots,\mathbf{c}_t]$, and the observation features are fused to the policy network
through FiLM \cite{perez2018film}.

Diffusion Policy adopts a DDPM \cite{ho2020denoising} denoising formulation. The forward diffusion process progressively corrupts a clean action
sequence by adding Gaussian noise:
\begin{equation}
q(\mathbf{A}_t^{n}\mid \mathbf{A}_t^{n-1})
= \mathcal{N}\!\left(\sqrt{\alpha_n}\mathbf{A}_t^{n-1},\ (1-\alpha_n)\mathbf{I}\right),
\end{equation}
where $\alpha_n\in(0,1)$, $\bar{\alpha}_n=\prod_{i=1}^{n}\alpha_i$ and $n\in\{1,\ldots,N\}$ denotes the diffusion step. Equivalently,
\begin{equation}
\mathbf{A}_t^{n} = \sqrt{\bar{\alpha}_n}\mathbf{A}_t^{0} + \sqrt{1-\bar{\alpha}_n}\,\boldsymbol{\epsilon},
\qquad \boldsymbol{\epsilon}\sim\mathcal{N}(\mathbf{0},\mathbf{I}).
\end{equation}
The reverse process is parameterized by a noise-prediction network
$\boldsymbol{\epsilon}_\theta(\mathbf{A}_t^{n},\mathbf{C}_t,n)$.
Let $\tilde{\mathbf{A}}_t^{n}=\sqrt{\bar{\alpha}_n}\mathbf{A}_t^{0}+\sqrt{1-\bar{\alpha}_n}\boldsymbol{\epsilon}$.
The denoiser is trained with the standard DDPM objective:
\begin{equation}
\mathcal{L}_{\mathrm{DP}}
=
\mathbb{E}_{t,n,\boldsymbol{\epsilon}}
\Big[
\big\|
\boldsymbol{\epsilon}
-
\boldsymbol{\epsilon}_\theta(\tilde{\mathbf{A}}_t^{n},\mathbf{C}_t,n)
\big\|_2^2
\Big].
\end{equation}
At inference, we use DDIM \cite{song2020denoising} sampling to efficiently generate an action sequence, and deploy it in a receding-horizon manner
by executing only the first $T_a$ actions ($T_a < T_p$) before replanning.

\subsection{Latency-Aware Classifier-Free Guidance}
CFG is a guidance technique for conditional diffusion models that strengthens adherence to a condition by combining
conditional and unconditional denoising predictions during sampling with a tunable guidance scale. 
In diffusion-based policies, CFG provides a lightweight way to incorporate a future-action condition to improve inter-chunk temporal consistency.
However, as illustrated in Fig.~\ref{fig:train}, existing future-action-conditioned CFG methods assume temporally aligned future-action conditions during training and can become brittle when inference latency shifts the effective condition at deployment~\cite{Arachchige2025SAIL}.

As illustrated in Fig.~\ref{fig:train}(a), the start time of the previous inference cycle is denoted as $t^{-}$. Under an asynchronous inference manner, the current inference cycle starts at
\begin{equation}
t \triangleq t^{-} + (T_a - T_d),
\end{equation}
where $T_d$ denotes the inference delay. There remains a segment of actions that were predicted in the previous cycle but not yet executed. As shown in the red box of Fig.~\ref{fig:train}(a), we extract a length-$T_f$ subsequence from this
unexecuted segment starting at $t$ and define it as the future-action condition for the current cycle: $\mathbf{A}^{c}_t = [\mathbf{a}_{t}, \ldots, \mathbf{a}_{t+T_f-1}]$.
When $\mathbf{A}^{c}_t$ is temporally aligned with the overlap between consecutive chunks, it provides an effective guidance signal that promotes cross-chunk continuity.

In SAIL~\cite{Arachchige2025SAIL}, the future-action condition is concatenated with the observation,
$\mathbf{o}_t=[\mathbf{v}_t,\mathbf{x}_t,\mathbf{A}^{c}_t]$, which yields the per-timestep conditioning vector
$\mathbf{c}_t=[\mathbf{z}_t,\mathbf{x}_t,\mathbf{A}^{c}_t]$. Consequently, the same future-action condition is replicated
across the $T_o$-step history conditioning vector $\mathbf{C}_t$, tightly coupling $\mathbf{A}^{c}_t$ with the history of
visual and proprioceptive features. 
Since $\mathbf{A}^{c}_t$ is meant to constrain the overlap between consecutive action chunks, this coupling makes the noise-prediction network $\boldsymbol{\epsilon}_\theta(\mathbf{A}_t^{n},\mathbf{C}_t,n)$ sensitive to temporal shifts in $\mathbf{A}^{c}_t$, which can degrade chunk-to-chunk continuity. 

To avoid this issue, we treat $\mathbf{A}^{c}_t$ as a separate conditioning variable and inject it through CFG during sampling.
Given a noisy action sequence $\mathbf{A}_t^{n}$ at diffusion step $n$ and the observation conditioning $\mathbf{C}_t$, our CFG computes conditional and unconditional noise predictions as
\begin{equation}
\left\{
\begin{aligned}
\boldsymbol{\epsilon}_{\mathrm{cond}} &:= \boldsymbol{\epsilon}_\theta(\mathbf{A}_t^{n},\, \mathbf{C}_t,\, \mathbf{A}^{c}_t,\, n),\\
\boldsymbol{\epsilon}_{\mathrm{uncond}} &:= \boldsymbol{\epsilon}_\theta(\mathbf{A}_t^{n},\, \mathbf{C}_t,\, \varnothing,\, n), \qquad \varnothing \equiv \mathbf{0},
\end{aligned}
\right.
\end{equation}
and combines them to obtain the guided noise estimate
\begin{equation}
\hat{\boldsymbol{\epsilon}}
=
\boldsymbol{\epsilon}_{\mathrm{uncond}}
+
w\big(\boldsymbol{\epsilon}_{\mathrm{cond}}-\boldsymbol{\epsilon}_{\mathrm{uncond}}\big),
\end{equation}
where $w$ is the guidance scale. In DDIM sampling, $\hat{\boldsymbol{\epsilon}}$ is used in place of the original noise
prediction at each denoising step, yielding a denoising trajectory that is more consistent with the future-action
condition $\mathbf{A}^{c}_t$.

\begin{algorithm}[b!]
\caption{Latency-Aware CFG Training}
\label{alg:latency_cfg_train_simple}
\KwIn{Demo dataset $\mathcal{D}$; horizons $T_o,T_p,T_f$; diffusion steps $n\sim\{1,\ldots,N\}$; drop prob $p$; shift distribution $\delta\sim\mathcal{P}_\delta$}
\For{each training iteration}{
    Sample timestep $t$ and extract $\mathbf{O}_t,\,\mathbf{A}_t^0=[\mathbf{a}_t,\ldots,\mathbf{a}_{t+T_p-1}]$\;
    Sample latency shift $\delta\sim\mathcal{P}_\delta$\;
    Set $\mathbf{A}_t^{c} = [\mathbf{a}_{t+\delta},\ldots,\mathbf{a}_{t+\delta+T_f-1}]$\;
    
    With probability $p$: $\mathbf{A}_t^{c}\leftarrow \mathbf{0}$\;
    
    Form noisy actions $\mathbf{A}_t^{n}=\sqrt{\bar{\alpha}_n}\mathbf{A}_t^{0}+\sqrt{1-\bar{\alpha}_n}\boldsymbol{\epsilon}$\;
    Update $\theta$ by minimizing $\|\boldsymbol{\epsilon}-\epsilon_\theta(\mathbf{A}_t^{n},\mathbf{C}_t,\mathbf{A}_t^{c},n)\|_2^2$\;
}
\end{algorithm}

In conventional training, the future-action condition is constructed to be temporally aligned with the target action chunk, as shown in Fig.~\ref{fig:train}(b).
However, under asynchronous inference, the time at which $\mathbf{A}^{c}_t$ is generated is decoupled from the time at which it is actually used for action execution. 
As shown in Fig.~\ref{fig:train}(c), inference latency can shift the effective $\mathbf{A}^{c}_t$ by $\delta$ action steps relative to the aligned training case, so the denoiser is guided to match the overlap at incorrect temporal positions, degrading inter-chunk continuity.
Training only with aligned future-action conditions therefore makes the policy brittle to such shifts.

To address this issue, we introduce delay randomization during training. Specifically, we sample an offset $\delta$ in action steps and shift $\mathbf{A}^{c}_t$ accordingly, as detailed in Algo.~\ref{alg:latency_cfg_train_simple}.
To train a single noise-prediction network for both conditional and unconditional denoising, we adopt the CFG dropout scheme~\cite{ho2022classifier}. 
This exposes the denoiser to shifted future-action conditions during training, improving robustness to future-action condition shifts and enabling smoother inter-chunk execution under asynchronous inference.

\subsection{Goal-Prediction Head for Diffusion Policy}
Prior safe diffusion-policy deployments remain within a receding-horizon control manner, applying safety corrections only to short-horizon action chunks. 
However, efficient and globally coherent obstacle avoidance requires an explicit task objective beyond short-horizon correction. 
Therefore, we equip the diffusion policy with a goal-prediction head that enables task-relevant goal prediction, which guides collision-free trajectory generation in cluttered scenes.

Specifically, we attach a goal-prediction head to the denoising U-Net~\cite{ronneberger2015u} to predict a task-relevant goal from its bottleneck representation $\mathbf{h}_t$, as shown in Fig.~\ref{fig:overview}.
The feature $\mathbf{h}_t$ is taken from the lowest-resolution stage of the U-Net encoder and serves as a compact latent summary that aggregates global context from the observation conditioning while capturing the coarse motion intent encoded in the denoised action chunk.
This combination makes $\mathbf{h}_t$ well suited for inferring task-relevant interaction goals.
Global average pooling (GAP) is applied to $\mathbf{h}_t$ to average over the action-horizon axis, yielding a fixed-length feature for goal prediction.
The goal $\hat{\mathbf{g}}_t$ is then predicted by a multi-layer perceptron (MLP):
\begin{equation}
\hat{\mathbf{g}}_t = g_\psi\!\left(\mathrm{GAP}(\mathbf{h}_t)\right).
\end{equation}
Goal supervision is imposed with the regression loss
\begin{equation}
\mathcal{L}_{\mathrm{goal}}(\theta,\psi)=\left\| \mathbf{g}_t^{*}-\hat{\mathbf{g}}_t \right\|_2^2,
\end{equation}
where $\theta$ denotes the parameters of Diffusion Policy including the observation encoders and denoising U-Net, $\psi$ denotes the parameters of the goal-prediction head, and $\mathbf{g}_t^{*}$ is a task-relevant goal extracted from expert demonstrations. The overall training objective is a
weighted sum of the diffusion denoising loss and the goal loss
\begin{equation}
\mathcal{L}(\theta,\psi)=\mathcal{L}_{\mathrm{DP}}(\theta)+\lambda_{\mathrm{goal}}\mathcal{L}_{\mathrm{goal}}(\theta,\psi),
\end{equation}
where nonnegative weight $\lambda_{\mathrm{goal}}$ balances noise prediction and goal prediction.

\subsection{Collision-Free Trajectory Generation}
After obtaining the task-relevant goal $\hat{\mathbf{g}}_t$, we generate a task-space end-effector motion toward $\hat{\mathbf{g}}_t$.
The end-effector is approximated as a collision sphere centered at the current position $\mathbf{x}^{\mathrm{ee}}_t$ with radius $r_{\mathrm{ee}}$.
We perform a line-of-sight collision check along the straight segment from $\mathbf{x}^{\mathrm{ee}}_t$ to $\hat{\mathbf{g}}_t$, testing whether the swept sphere intersects any obstacles, and compute the goal distance $d_t=\|\mathbf{x}^{\mathrm{ee}}_t-\hat{\mathbf{g}}_t\|_2$.
Trajectory optimization is triggered if a collision is detected or if $d_t$ exceeds a threshold $d_{\mathrm{th}}$.
When the line-of-sight motion is collision-free, the optimizer generates a smooth and feasible goal-reaching trajectory; when obstacles obstruct the motion, the collision term enforces obstacle clearance.

The avoidance module draws inspiration from EGO-Planner~\cite{zhou2020ego}, an efficient spline-based optimizer that does not require a collision-free initialization.
This property aligns with our setting, where a natural initialization is the direct motion toward $\hat{\mathbf{g}}_t$, which may intersect obstacles.
Specifically, the end-effector position trajectory is parameterized by a uniform B-spline curve $\Phi(t)$ whose decision
variables are the control points $\mathbf{Q}=\{\mathbf{Q}_i\}_{i=1}^{N_c}$:
\begin{equation}
\label{eqn:b-spline}
\Phi(t)=\sum_{i=1}^{N_c} B_{i,d}(t)\,\mathbf{Q}_i,
\end{equation}
where $N_c$ is the number of control points and $B_{i,d}(t)$ denotes the B-spline basis of degree $d$. 

\begin{figure}[t!]
	\vspace{3mm}
    \centering
        \includegraphics[width=0.7\linewidth]{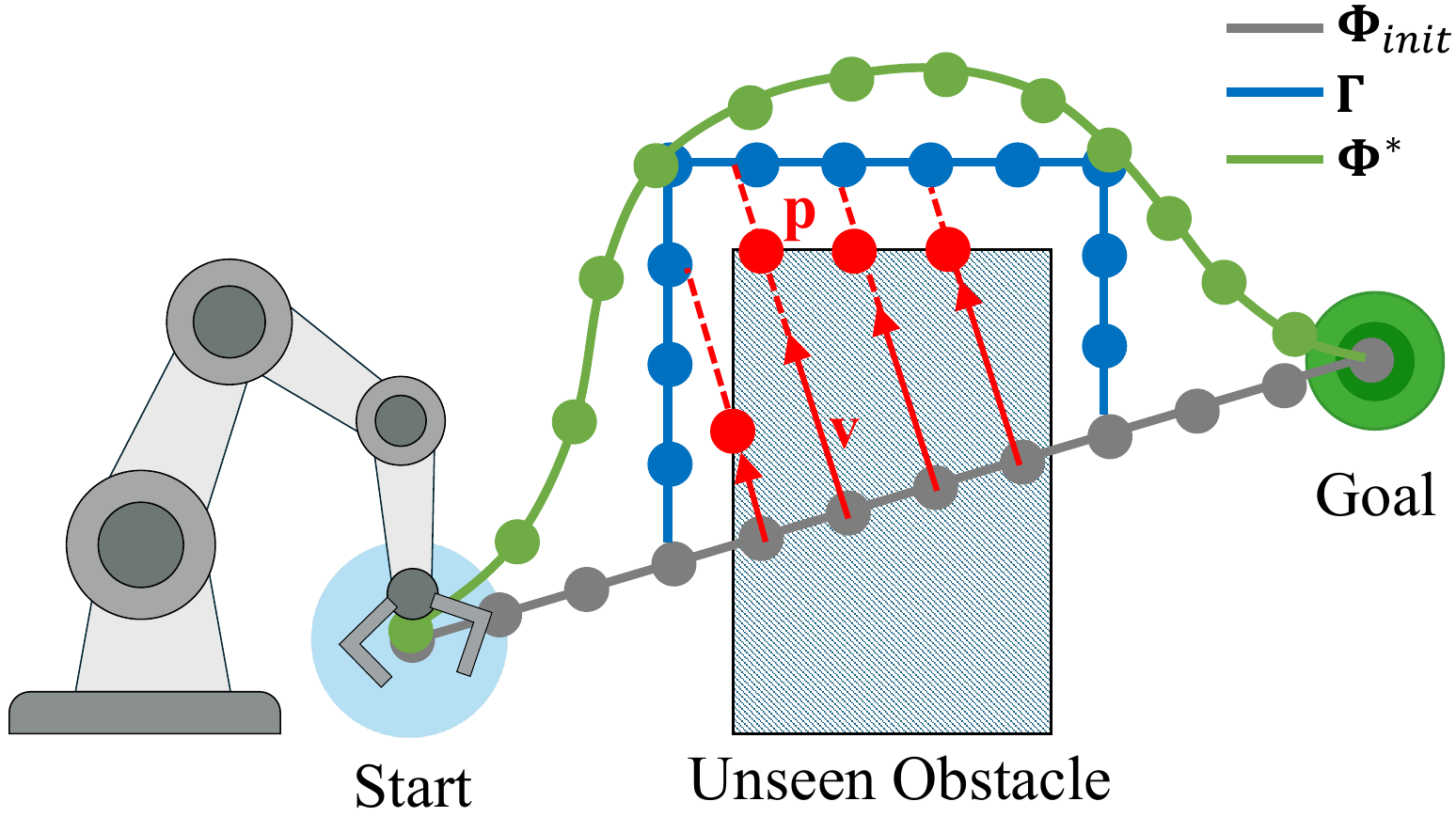}	
        \caption{The initial trajectory $\Phi_{\mathrm{init}}$ collides with an unseen obstacle. A* provides a collision-free connector $\Gamma$, from which $\{\mathbf{p},\mathbf{v}\}$ pairs guide optimization to a smooth collision-free trajectory $\Phi^{*}$.}
	    \label{fig:comparison}
        \vspace{-5mm}
\end{figure}

As shown in Fig.~\ref{fig:comparison}, an initial trajectory $\Phi_{\mathrm{init}}$ is constructed to satisfy terminal constraints between the current end-effector position $\mathbf{x}^{\mathrm{ee}}_t$ and the predicted goal $\hat{\mathbf{g}}_t$, without requiring collision-free initialization.
Trajectory generation is formulated as minimizing a weighted sum of smoothness, collision, and feasibility costs:
\begin{equation}
\label{eqn:ego_objective}
\arg\min_{\mathbf{Q}}\; \mathcal{J}(\mathbf{Q})
=
\lambda_s \mathcal{J}_{\mathrm{s}}(\mathbf{Q})
+
\lambda_c \mathcal{J}_{\mathrm{c}}(\mathbf{Q})
+
\lambda_f \mathcal{J}_{\mathrm{f}}(\mathbf{Q}),
\end{equation}
where $\mathcal{J}_{\mathrm{s}}$ penalizes high-order derivatives for smooth motion, $\mathcal{J}_{\mathrm{f}}$ penalizes
violations of physical limits, and $\mathcal{J}_{\mathrm{c}}$ enforces obstacle clearance. 
The nonnegative weights $\lambda_s$, $\lambda_c$, and $\lambda_f$ trade off smoothness, collision avoidance, and feasibility, respectively.
The specific forms of $\mathcal{J}_{\mathrm{s}}$ and $\mathcal{J}_{\mathrm{f}}$ follow standard spline-optimization formulations and are referred to EGO-Planner~\cite{zhou2020ego}. 
The collision cost is detailed below.

For each colliding segment detected on $\Phi_{\mathrm{init}}$, a collision-free guiding path $\mathbf{\Gamma}$ is generated using
A* \cite{hart1968formal}. Each control point $\mathbf{Q}_i$ on the colliding segment is assigned $N_p(i)$ anchor-direction pairs
$\{\mathbf{p}_{ij},\mathbf{v}_{ij}\}_{j=1}^{N_p(i)}$, where $i\in\mathbb{N}_{+}$ indexes control points and
$j\in\{1,\ldots,N_p(i)\}$ indexes the pairs associated with $\mathbf{Q}_i$. For each pair, $\mathbf{v}_{ij}$ is a unit
repulsive direction that is orthogonal to the local spline tangent at $\mathbf{Q}_i$ and oriented toward the guiding path
$\mathbf{\Gamma}$. The anchor point $\mathbf{p}_{ij}$ is selected on the obstacle surface along the direction of $\mathbf{v}_{ij}$, forming a signed distance 
$d_{ij} = (\mathbf{Q}_i - \mathbf{p}_{ij})^\top \mathbf{v}_{ij}$.
A safety clearance is enforced by adopting a margin $s_m$ beyond the end-effector sphere radius $r_{\mathrm{ee}}$. For each
anchor-direction pair, we define the clearance violation relative to the signed distance $d_{ij}$ and apply a twice
continuously differentiable penalty:
\begin{equation}
\begin{aligned}
\phi(c_{ij}) &=
\begin{cases}
0, & c_{ij}\le 0,\\
c_{ij}^3, & 0<c_{ij}\le s_f,\\
3s_f c_{ij}^2-3s_f^2 c_{ij}+s_f^3, & c_{ij}> s_f,
\end{cases} \\
c_{ij} &= s_f-d_{ij}, \qquad s_f=r_{\mathrm{ee}}+s_m .
\end{aligned}
\end{equation}
The collision cost for each control point $\mathbf{Q}_i$ is accumulated over its associated pairs, and the overall
collision cost is
\begin{equation}
\mathcal{J}_{\mathrm{c}}(\mathbf{Q})
=
\sum_{i=1}^{N_c}\sum_{j=1}^{N_p(i)} \phi(c_{ij}).
\end{equation}
The collision term induces gradients colinear with $\mathbf{v}_{ij}$, deforming colliding control points along these directions to push the spline out of obstacles. 
The optimized trajectory $\Phi^*(t)$ is then obtained by substituting the optimized control points $\mathbf{Q}^*$ from~(\ref{eqn:ego_objective}) into~(\ref{eqn:b-spline}).

\subsection{Spatial-Temporal Trajectory Optimization}
The discrete action sequence $\mathbf{A}_t$ can be obtained either from the diffusion policy or by sampling the collision-free trajectory $\Phi^*(t)$. It is treated as a set of action keypoints, and is refined by spatial-temporal trajectory optimization to reduce control effort while enforcing feasibility constraints.

Let $\Delta t$ denote the sampling period and define the trajectory time allocation as $T = T_a \Delta t$.
A continuous-time action trajectory is parameterized as polynomial
$\mathbf{A}(t)=\mathbf{c}^\top \boldsymbol{\beta}(t)$, where $\boldsymbol{\beta}(t)=[1,t,\ldots,t^{2s-1}]^\top$ is the time
basis and $\mathbf{c}$ is a coefficient matrix.
The refinement is formulated as
\begin{equation}
\begin{aligned}
\min_{\mathbf{c}}\;\; 
J(\mathbf{c})
&=
\lambda_E J_E
+
\lambda_T J_T
+
\lambda_P J_P \\
\text{s.t.}\;\;
\mathbf{A}(k\Delta t) &= \mathbf{a}_{t+k}, \qquad k=0,\ldots,T_a-1,
\end{aligned}
\label{eq:st_opt}
\end{equation}
where $\lambda_E$, $\lambda_T$, and $\lambda_P$ are nonnegative weights that balance control effort, execution time, and
feasibility, respectively. The control-effort term is
\begin{equation}
J_E = \int_{0}^{T}\left\|\mathbf{A}^{(s)}(t)\right\|_2^2\,dt,
\end{equation}
the time regularization term is $J_T = T$, and $J_P$ penalizes violations of action limits over $t\in[0,T]$, including
elementwise bounds on action, velocity, and acceleration:
\begin{equation}
\begin{aligned}
\mathbf{A}_{\min} \le \mathbf{A}(t) \le \mathbf{A}_{\max},\\
\dot{\mathbf{A}}_{\min} \le \dot{\mathbf{A}}(t) \le \dot{\mathbf{A}}_{\max},\\
\ddot{\mathbf{A}}_{\min} \le \ddot{\mathbf{A}}(t) \le \ddot{\mathbf{A}}_{\max}.
\end{aligned}
\end{equation}
In practice, these inequality constraints are transcribed into a differentiable penalty and evaluated along the trajectory. 
The optimization is solved efficiently using MINCO \cite{wang2022geometrically}, and the optimized trajectory is resampled for execution.
\section{Experiments}
\label{sec:experiments}

We design experiments to answer the following questions:
\begin{itemize}
\item \textbf{Q1}: Does latency-aware classifier-free guidance maintain inter-chunk consistency under temporally shifted future-action conditions?
\item \textbf{Q2}: Does LAGO Policy produce smoother execution trajectories in real-world manipulation?
\item \textbf{Q3}: Does goal-directed collision-free planning improve task completion in the presence of unseen obstacles?
\end{itemize}
\subsection{Experiment Setup}

\begin{table*}[t!]
\vspace{3mm}
\centering
\caption{Real-World Evaluation}
\label{tab:real-eval_combined}
\scriptsize

\begin{tblr}{
  width       = 0.97\textwidth,
  colsep      = 2pt,
  colspec = {Q[l] | XXXX | XXXX | XXXX | XXXX },
  row{1}      = {font=\bfseries},
  hline{1,3,5}= {-}{},
  cell{1}{2}  = {c=4}{c},
  cell{1}{6}  = {c=4}{c},
  cell{1}{10} = {c=4}{c},
  cell{1}{14} = {c=4}{c},
}
 & Pick \& Place & & & &
   Pen Insertion  & & & &
   Pouring        & & & &
   Cup Transfer & & & \\
Method &
SR$\uparrow$ & ATR$\downarrow$ & CON$\downarrow$ & ISJ$\downarrow$ &
SR$\uparrow$ & ATR$\downarrow$ & CON$\downarrow$ & ISJ$\downarrow$ &
SR$\uparrow$ & ATR$\downarrow$ & CON$\downarrow$ & ISJ$\downarrow$ &
SR$\uparrow$ & ATR$\downarrow$ & CON$\downarrow$ & ISJ$\downarrow$ \\

DP &
 0.85 & 16.91 & 0.034 & 27.29 &
 0.70 & 8.76 & \textbf{0.0092} & 32.71 &
 0.25 & 20.24 & 0.033 & 22.44 &
 0.10 & \textbf{10.27} & 0.041 & 26.34 & \\

\textbf{Ours} &
 \textbf{1.0}  & \textbf{15.41} & \textbf{0.019} & \textbf{24.37} &
 \textbf{0.90} & \textbf{7.94}  & 0.011 & \textbf{21.45} &
 \textbf{0.65} & \textbf{19.41} & \textbf{0.012} & \textbf{15.74} &
 \textbf{0.40} & 10.62 & \textbf{0.0083} & \textbf{10.11} & \\
\end{tblr}


\begin{tblr}{
  width       = 0.97\textwidth,
  colsep      = 2pt,
  colspec = {Q[l] | XXXX | XXXX | XXXX | XXXX },
  row{1}      = {font=\bfseries},
  hline{1,3,5}= {-}{},
  cell{1}{2}  = {c=4}{c},
  cell{1}{6}  = {c=4}{c},
  cell{1}{10} = {c=4}{c},
  cell{1}{14}  = {c=4}{c},
}
 & Towel Folding & & & &
   Box Organizing & & & &
   Tape Hanging & & & &
   Screw Sorting & & & \\

Method &
SR$\uparrow$ & ATR$\downarrow$ & CON$\downarrow$ & ISJ$\downarrow$ &
SR$\uparrow$ & ATR$\downarrow$ & CON$\downarrow$ & ISJ$\downarrow$ &
SR$\uparrow$ & ATR$\downarrow$ & CON$\downarrow$ & ISJ$\downarrow$ &
SR$\uparrow$ & ATR$\downarrow$ & CON$\downarrow$ & ISJ$\downarrow$ \\

DP &
 0.35 & 18.24 & 0.058 & 33.11 &
 0.60 & 28.13 & 0.072 & 96.10 &
 0.80 & \textbf{32.82} & 0.0066 & 115.51 &
 0.60 & 53.50 & 0.0062 & 116.67 & \\

\textbf{Ours} &
 \textbf{0.70} & \textbf{14.27} & \textbf{0.017} & \textbf{15.25} &
 \textbf{0.85} & \textbf{27.31} & \textbf{0.026} & \textbf{29.07} &
 \textbf{0.90} & 36.80 & \textbf{0.0059} & \textbf{21.16} &
 \textbf{0.95} & \textbf{52.97} & \textbf{0.0037} & \textbf{18.77} & \\
\end{tblr}
\vspace{-1mm}
\end{table*}

\begin{figure*}[t!]
  \centering
  \includegraphics[width=0.97\textwidth]{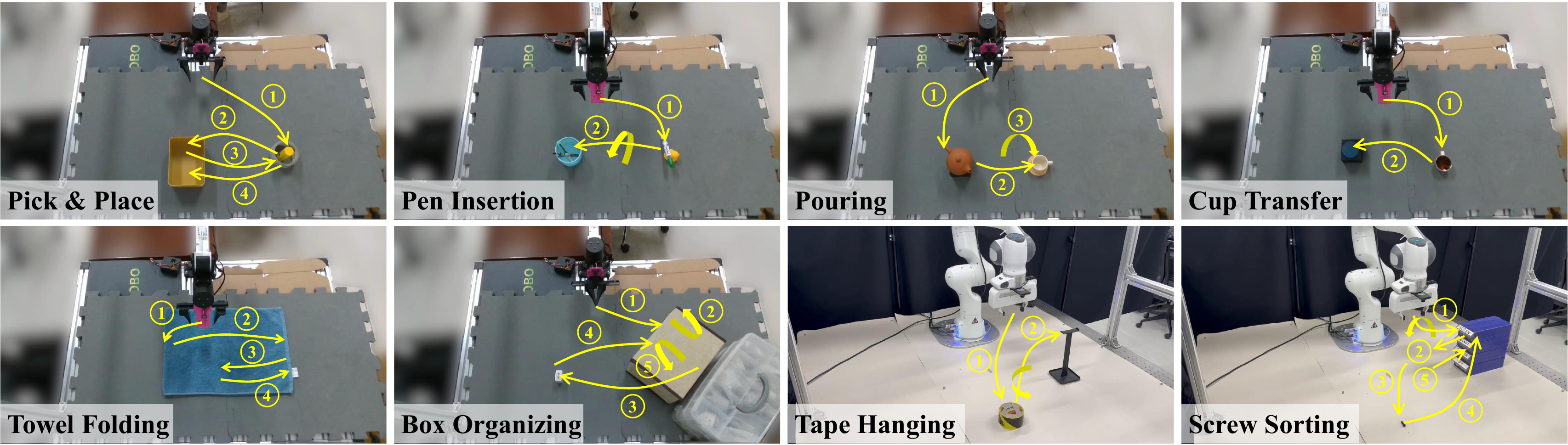}
  \caption{Real-world task setups. \textit{Tape Hanging} and \textit{Screw Sorting} are evaluated on the Franka arm, while the remaining tasks are evaluated on ARX5.}
  \label{fig:tasks}
\vspace{-5mm}
\end{figure*}

\subsubsection{Hardware Platform} 
Experiments are conducted on two manipulation platforms: a 6-DoF ARX5 arm and a 7-DoF Franka arm. 
Each setup is equipped with two external Intel RealSense D435 / L515 RGB-D cameras that provide workspace-scale coverage of the manipulation area, and a wrist-mounted fisheye camera with a 180$^\circ$ field of view that captures
close-up observations for fine manipulation.  
All devices are connected to a workstation with an Intel Core i7-14700 CPU and an NVIDIA RTX 4070 SUPER GPU, which is used for both data collection and policy evaluation.

\subsubsection{Task}
As illustrated in Fig.~\ref{fig:tasks}, we evaluate our method on eight real-world manipulation tasks spanning pick-and-place, insertion, transport of liquid-filled objects, deformable object manipulation, and articulated-object manipulation.
These tasks require accurate grasping and end-effector pose alignment, stable contact-rich control, and smooth low-jerk execution.
For example, \textit{Cup Transfer} requires the robot to transport a tea-filled cup and place it onto a coaster without spilling, while \textit{Pouring} requires the gripper to precisely thread through the teapot handle and maintain stable motion during pouring.
Additional task details and full execution sequences are provided in the supplementary video.

\subsubsection{Expert Demonstrations}
Expert demonstrations for all tasks are collected via human teleoperation.
The ARX5 arm is controlled using a SpaceMouse, whereas the Franka arm is controlled via keyboard teleoperation.
For each task, we record 50 demonstrations, each consisting of multi-view RGB images, end-effector poses, gripper states, and teleoperation commands.

\subsubsection{Implementation Details}
We follow the Diffusion Policy setup with an action prediction horizon $T_p=16$, an observation horizon $T_o=2$, and an
action execution horizon $T_a=8$. 
Following SAIL, we set the future-action horizon to $T_f=4$. 
We use a DDIM noise scheduler with 100 diffusion steps during training and 8 denoising steps at inference. Models are optimized with AdamW using an initial learning rate of $1\times10^{-4}$ and a cosine decay schedule. 
We train the model for 350 epochs with a batch size of 64 for each task.

\subsubsection{Evaluation Metrics}
Our evaluation uses metrics that capture both task-level performance and motion quality.
Task performance and efficiency are measured by the success rate (\textbf{SR}) and the average time of successful rollouts (\textbf{ATR}).
Motion quality is evaluated by inter-chunk consistency (\textbf{CON}) and trajectory smoothness. 
$\textbf{CON} = ||\mathbf{A}^{c}_t-\mathbf{A}_t^{0:T_f-1}||$ measures the discrepancy between the overlapping parts of consecutive action chunks at the chunk boundary, where $\mathbf{A}_t^{0:T_f-1}$ denotes the first $T_f$ actions of the current chunk $\mathbf{A}_t$.
Smoothness is quantified by the integrated squared jerk (\textbf{ISJ}) of the executed action trajectory:
$\textbf{ISJ} = \int_{0}^{T_{\mathrm{exec}}} ||\dddot{\mathbf{A}}(t)||^2_2\,dt$, where $T_{\mathrm{exec}}$ is the execution duration of the rollout. 

\subsection{Robustness to Future-Action Condition Shifts}
\textbf{Latency-aware CFG is robust to temporally shifted future-action conditions (Q1).}
We evaluate robustness to future-action condition shifts, which can arise from asynchronous inference latency and tracking error, by comparing three variants: (i) \textsc{SAIL}, which concatenates $\mathbf{A}^{c}_t$ with the observation and replicates it across the
$T_o$-step conditioning sequence; (ii) \textsc{Ours-CFG}, which injects $\mathbf{A}^{c}_t$ as a separate conditioning
variable via CFG during sampling but trains without delay randomization; and (iii) \textsc{Ours-LA-CFG}, which
additionally randomizes the temporal offset of $\mathbf{A}^{c}_t$ during training. Experiments are conducted on \textit{Box Organizing} task, which is sensitive to inter-chunk continuity. 
We set the guidance scale to $w=1$, run 20 rollouts per setting, and report \textbf{SR} and \textbf{CON} under simulated future-action condition shifts of 0--4 steps.

\begin{figure}[t!]
    \vspace{3mm}
	\centering        \includegraphics[width=0.95\linewidth]{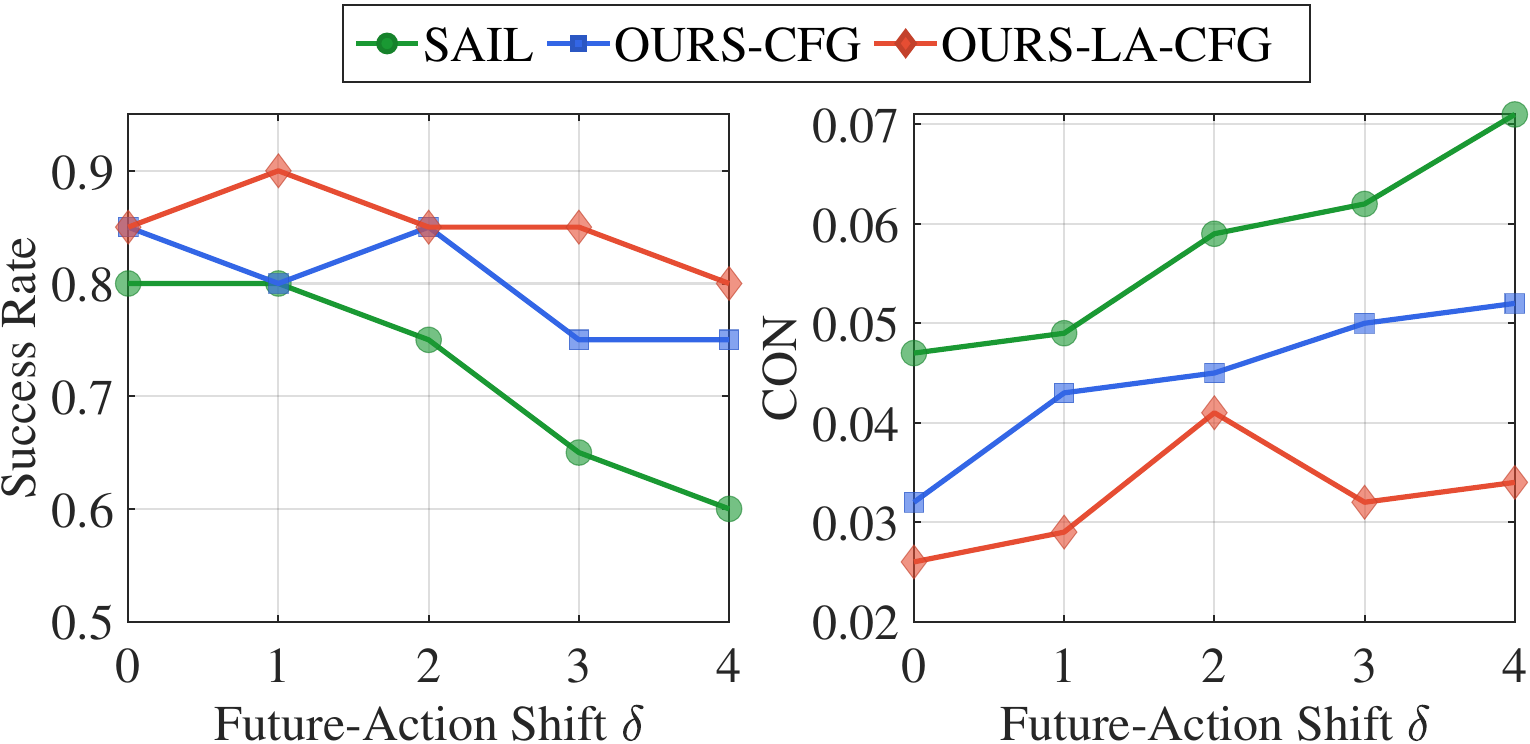}	
        \caption{\textbf{SR} and \textbf{CON} under simulated future-action shifts $\delta$. As the shift increases, OURS-LA-CFG maintains a higher success rate and exhibits less inter-chunk discontinuity.}
	    \label{fig:cfg_shift_ablation}
\end{figure}

The results are shown in Fig.~\ref{fig:cfg_shift_ablation}. 
We first compare \textsc{SAIL} and \textsc{Ours-CFG}. 
As the future-action condition becomes increasingly shifted, both methods exhibit degraded task performance and motion quality, with \textbf{SR} decreasing and \textbf{CON} increasing. 
However, \textsc{Ours-CFG} achieves higher \textbf{SR} and a smaller increase in \textbf{CON} than \textsc{SAIL}, indicating improved robustness and better
inter-chunk continuity under misaligned future-action conditions. 
This benefit is attributed to the conditioning interface. In \textsc{SAIL}, the shifted future-action condition is embedded into the observation history, leading to repeated exposure to a misaligned signal across $T_o$ timesteps.
In contrast, \textsc{Ours-CFG} injects $\mathbf{A}^{c}_t$ only once through CFG during sampling, which mitigates the
adverse effect of temporally shifted conditioning on denoising and action generation.
We then compare \textsc{Ours-CFG} and \textsc{Ours-LA-CFG}. 
In contrast to the monotonic degradation observed above, \textsc{Ours-LA-CFG} remains markedly more stable as the shift increases, maintaining a higher \textbf{SR} and exhibiting a smaller change in \textbf{CON}. 
These results show that delay randomization during training further improves robustness to temporally shifted future-action conditions at deployment, yielding more consistent inter-chunk execution.

\subsection{Smooth Execution in Real-World Manipulation}
\textbf{LAGO Policy produces smoother real-world execution (Q2).}
We compare motion quality between Diffusion Policy and LAGO Policy with its goal-directed planning module disabled on eight real-world manipulation tasks, as shown in Fig.~\ref{fig:tasks}.
We deploy the trained models for 20 rollouts per task and log robot states during execution. The results in Table~\ref{tab:real-eval_combined} show that LAGO Policy achieves lower \textbf{CON} on seven tasks and lower \textbf{ISJ} on all tasks, indicating improved inter-chunk continuity and smoother execution trajectories.

Qualitatively, LAGO Policy is more robust to action ambiguity induced by similar observations across consecutive cycles. 
For example, in \textit{Screw Sorting}, opening and closing the sliding drawer often yields near-identical visual states, under which Diffusion Policy may hesitate between modes and produce stop-and-go motions. 
Conditioning with future-action guidance encourages temporally consistent chunk transitions and yields continuous motion.
Moreover, for contact-rich tasks where stable gripper behavior is critical, such as \textit{Towel Folding} and \textit{Pouring}, we observe that Diffusion Policy can generate inconsistent gripper commands after grasping, occasionally opening the gripper during transport and causing object drops or spills. 
LAGO Policy improves gripper command continuity, leading to more reliable contact maintenance and higher rollout success.

To further assess the contribution of spatial-temporal trajectory optimization to motion quality, we conduct an ablation on \textit{Box Organizing} by removing the optimization module from LAGO Policy. 
Without this module, \textbf{CON} increases to $0.041$ and \textbf{ISJ} increases to $40.03$, indicating degraded inter-chunk continuity and higher jerk in the executed trajectory. 
These results suggest that spatial-temporal trajectory optimization improves trajectory smoothness, which is particularly beneficial for tasks requiring low-jerk motions.

\begin{figure}[t!]
    \vspace{3mm}
	\centering
        \includegraphics[width=0.9\linewidth]{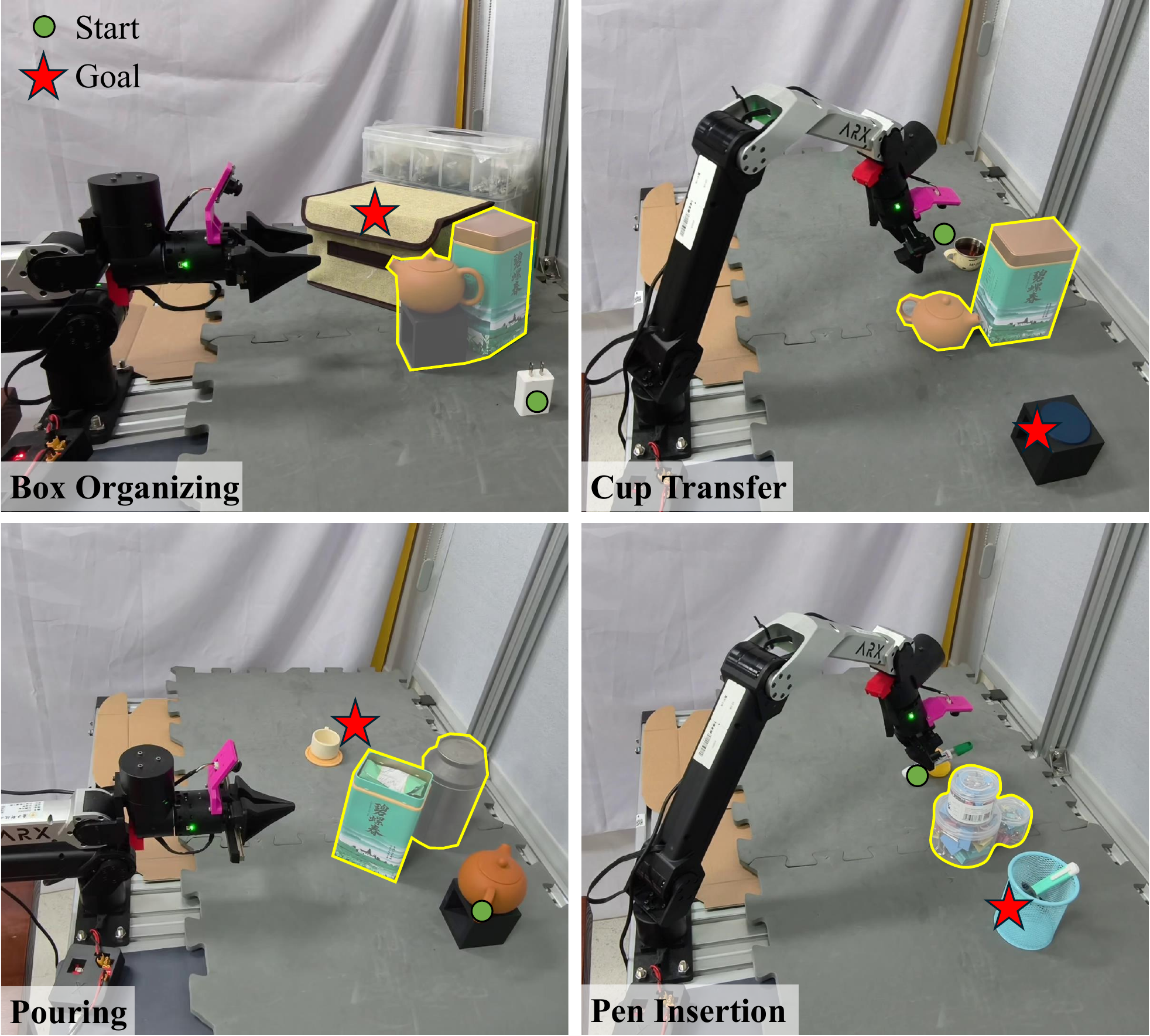}	
        \caption{Obstacles absent from the training demonstrations are added to four tasks and highlighted in yellow. Our \textsc{Goal-Directed} method plans smooth collision-free motion from the start to the predicted goal.}
	    \label{fig:obstacle_avoidance}
\end{figure}

\begin{table}[b!]
    \centering
    \caption{Success rate on four real-world manipulation tasks.}
    \label{tab:sr_four_tasks}
    \setlength{\tabcolsep}{6pt}
    \renewcommand{\arraystretch}{1.15}
    \begin{tabular}{ccccc}
        \toprule
        \textbf{Method} & \textbf{Box} & \textbf{Cup} & \textbf{Pouring} & \textbf{Pen} \\
        \midrule
        \textsc{NO-Avoid}         & 0.40  & 0.0  & 0.0  & 0.60 \\
        \textsc{Local}            & 0.60  & 0.20 & 0.40 & 0.80 \\
        \textsc{Goal-Directed}    & \textbf{1.0}   & \textbf{0.80} & \textbf{0.80} & \textbf{1.0} \\
        \bottomrule
    \end{tabular}
\end{table}

\subsection{Goal-Directed Obstacle Avoidance and Task Success}
\textbf{Goal-directed collision-free planning improves task completion under unseen obstacles (Q3).}
As shown in Fig.~\ref{fig:obstacle_avoidance}, we introduce unseen obstacles that are absent from the expert demonstrations in four tasks and evaluate task success rate by comparing three variants:
(i) \textsc{No-Avoid}, which disables the collision-free trajectory generation module and directly executes the diffusion policy outputs;
(ii) \textsc{Local}, which replaces collision-free trajectory generation with a local safety filter; and
(iii) \textsc{Goal-Directed}, our full method with goal prediction and goal-conditioned collision-free planning.
The \textsc{Local} baseline performs collision checking on the action chunk predicted by diffusion policy at each inference cycle. 
When a collision is detected, the colliding actions are projected to the nearest collision-free end-effector states in task space, and A* is used to generate a short collision-free detour for execution.
This setup follows a common local safety filtering paradigm that enforces collision avoidance by minimally modifying the policy output \cite{hu2025vlsa}.

Each method is evaluated with five independent rollouts per task, and the resulting success rates are reported in Table~\ref{tab:sr_four_tasks}.
Across all tasks, \textsc{No-Avoid} frequently collides with the introduced obstacles, leading to low success rates.
This effect is particularly pronounced in \textit{Cup Transfer}, where even slight contact perturbs the liquid and causes failure.
While \textsc{Local} reduces direct collisions, its short-horizon corrections often push the robot into out-of-distribution states relative to the demonstrations, inducing unstable behaviors such as abrupt gripper commands and jerk motions that hinder task completion.
For example, in \textit{Pouring}, an upward reactive avoidance maneuver induces jerky corrections and an unintended gripper opening, causing the teapot to tip over.
In contrast, \textsc{Goal-Directed} generates a smooth collision-free trajectory toward the predicted task goal and then hands control back to the policy for the final interaction, thereby avoiding the failure modes induced by reactive short-horizon corrections in \textsc{Local} and achieving the highest success rates under unseen obstacles.

\subsection{Goal-Directed Planning for Long-Range Motion}
We further study the effect of goal-directed planning on long-range motion efficiency.
On Franka, demonstrations are collected via keyboard teleoperation, which often produces axis-aligned, piecewise motions.
The learned policy consequently reproduces such trajectories, resulting in long travel distance and increased task completion time.
Goal-directed planning instead generates a smooth direct trajectory toward the task-relevant goal, substantially reducing task completion time, as shown in Fig.~\ref{fig:franka_goal} and in the supplementary video.

\begin{figure}[t!]
    \vspace{3mm}
	\centering
    \includegraphics[width=0.97\linewidth]{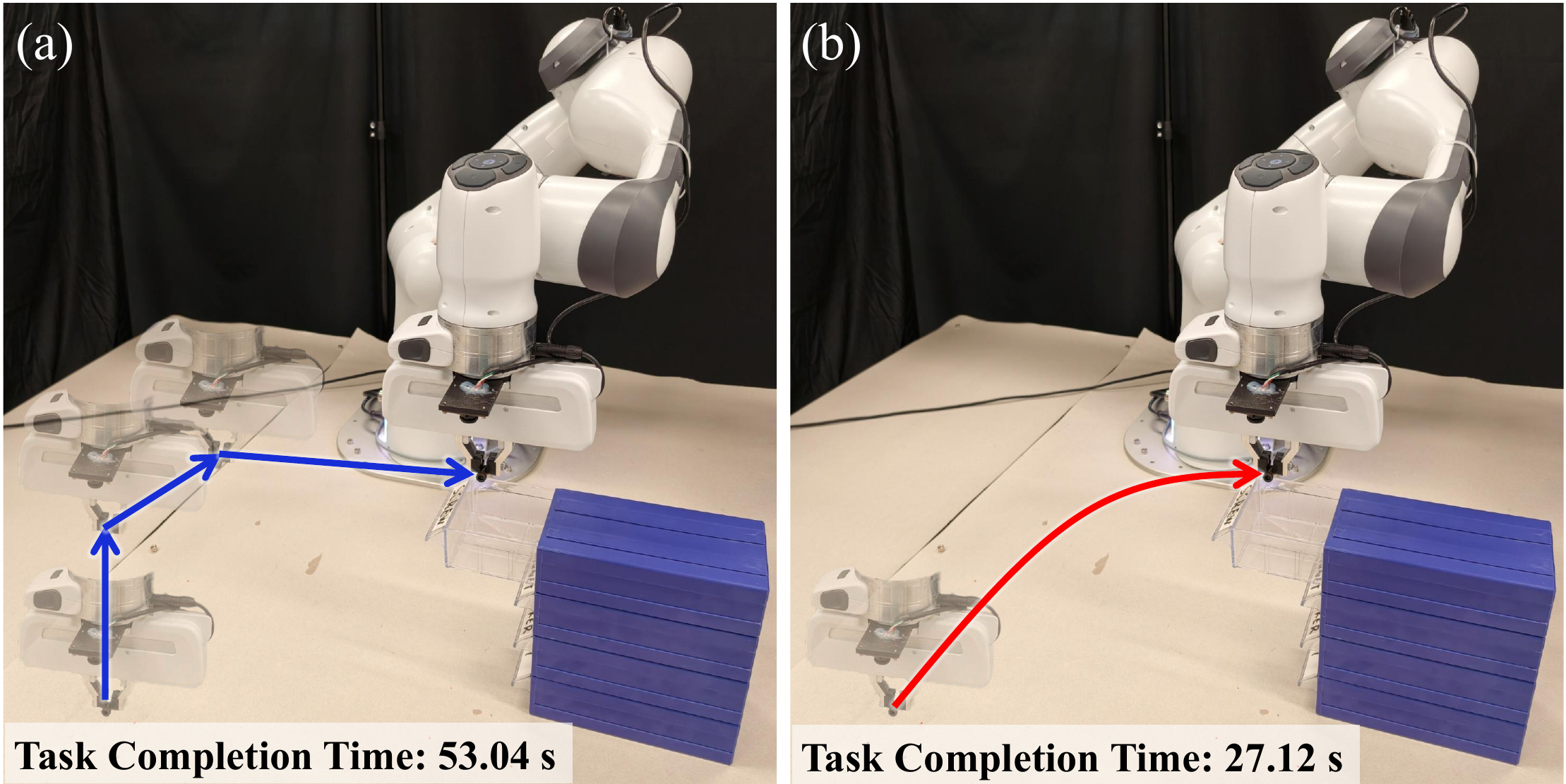}	
        \caption{Effect of goal-directed motion planning. (a) Without goal-directed motion planning, the motion remains piecewise and inefficient over long distances. (b) With goal-directed motion planning, the robot reaches the task-relevant goal along a smooth direct trajectory.}
	    \label{fig:franka_goal}
        \vspace{-2mm}
\end{figure}
\section{Conclusion}
\label{sec:conclusion}
We presented LAGO Policy, a unified framework for smooth and safe diffusion-based manipulation under asynchronous inference.
By combining latency-aware classifier-free guidance, goal-directed trajectory generation, and spatial-temporal trajectory optimization, LAGO Policy improves inter-chunk consistency, reduces jerk, and enables collision-free execution toward task-relevant goals.
Experiments on real-world manipulation tasks demonstrate smoother execution and improved task success in cluttered environments.
For future work, we will extend the current task-space collision handling to whole-body collision-aware motion planning, and improve task-relevant goal prediction in more complex manipulation settings.

\bibliographystyle{IEEEtran}
\bibliography{IEEEabrv,ref}

\end{document}